%% file: swdm-2016.tex
\documentclass{sig-alternate-05-2015}

\usepackage{times}
\usepackage{url}
\usepackage{latexsym}

\usepackage{caption}
\usepackage{graphicx}
\usepackage{color}
\usepackage{amsmath}
\usepackage{amssymb} 
\usepackage{booktabs}
\usepackage[normalem]{ulem}
\usepackage{comment}
\usepackage{multirow}
\usepackage{subfigure}
\usepackage{balance}

\usepackage{graphicx}
\usepackage{tikz}
\usetikzlibrary{plotmarks}
\usepackage{csvsimple,booktabs}
\usepackage{pgfplotstable}
\usepackage{pgfplots}
\usepackage{booktabs}
\usepackage{colortbl}
\usepackage{float}
\usepackage{algorithmic}
\usepackage[ruled,vlined]{algorithm2e}

\PassOptionsToPackage{hyphens}{url}
\usepackage{hyperref}

\usepackage{float}
\restylefloat{table}
\input{defs}

\usepackage{placeins}

\begin{document}






%

\title{Applications of Online Deep Learning for Crisis Response Using Social Media Information}

%
%
%
%
%

\numberofauthors{4} 
%
\author{
%
%
Dat Tien Nguyen, Shafiq Joty, Muhammad Imran, Hassan Sajjad, Prasenjit Mitra\\
       \affaddr{Qatar Computing Research Institute, HBKU}\\
       \affaddr{Doha, Qatar}\\
       \email{\{ndat, sjoty, mimran, hsajjad, pmitra\}@qf.org.qa}
}
\permission{This article has been peer reviewed and accepted at the 4th international workshop on Social Web for Disaster Management (SWDM), co-located with the 25th Conference of Information and Knowledge Management (CIKM), October 24--28, 2016, Indianapolis, USA.}
\copyrightetc{Copyright is held by the authors.}

\maketitle
\vspace{-3cm}
\begin{abstract}
During natural or man-made disasters, humanitarian response organizations look for useful information to support their decision-making processes. Social media platforms such as Twitter have been considered as a vital source of useful information for disaster response and management. Despite advances in natural language processing techniques, processing  short and informal Twitter messages is a challenging task. In this paper, we propose to use Deep Neural Network (DNN) to address two types of information needs of response organizations: \Ni identifying informative tweets and \Nii classifying them into topical classes. DNNs use distributed representation of words and learn the representation as well as higher level features automatically for the classification task. We propose a new online algorithm based on stochastic gradient descent to train DNNs in an online fashion during disaster situations. We test our models using a crisis-related real-world Twitter dataset. 
\end{abstract}

%
%

%
%


\keywords{deep learning, supervised classification, twitter, text classification, crisis response}

\section{Introduction}
Emergency events such as natural or man-made disasters bring unique challenges for humanitarian response organizations. Particularly, sudden-onset crisis situations demand officials to make fast decisions based on minimum information available to deploy rapid crisis response. However, information scarcity during time-critical situations hinders decision-making processes and delays response efforts~\cite{castillo2016big,imran2015processing}.

During crises, people post updates regarding their statuses, ask for help and other useful information, report infrastructure damages, injured people, etc.,
on social media platforms like Twitter~\cite{vieweg2014integrating}.
Humanitarian organizations can use this citizen-generated information to provide 
relief if critical information is easily available in a timely fashion.\footnote{\url{http://www.napsgfoundation.org/wp-content/uploads/2013/02/NAPSG-Remote-Sensing-Webcast-022213.pdf}} 
In this paper, we consider the classification of the social media posts into different humanitarian categories to fulfill different information needs of humanitarian organizations. Specifically, we address two types of information needs described as follows:

\smallskip
\noindent{\bf Informativeness of social media posts:}
Information posted on social networks during crises vary greatly in value. Most messages contain irrelevant information 
not useful for disaster response and management. 
Humanitarian organizations do not want a deluge of noisy messages that are of a personal nature or those that do not contain any useful information. 
They want clean data that consists of messages containing potentially useful information.
They can then use this information for various purposes such as situational awareness. 
In order to assist humanitarian organizations, we perform {\it binary classification}.
That is, we aim to classify each message into one of the two classes i.e. \textit{``informative"} vs. \textit{``not informative"}.

\smallskip
\noindent{\bf Information types of social media posts}
Furthermore, 
humanitarian organizations are interested in sorting social media posts into different categories. Identifying social media posts by category assists humanitarian organizations in coordinating their response. Categories such as infrastructure damage, reports of deceased or injured, urgent need for shelter, food and water, or donations of goods or services could therefore be directed to different relief functions.
%
In this work, we show how we can classify tweets into multiple classes.

Automatic classification of short crisis-related messages such as tweets is a challenging task due to a number of reasons. Tweets are short (only 140 characters), informal, often contain abbreviations, spelling variations and mistakes, 
and, therefore, 
they are hard to understand without enough context. Despite advances in natural language processing (NLP), interpreting the semantics of short informal texts automatically remains a hard problem. 
Traditional classification approaches rely on manually engineered features like cue words and TF-IDF vectors for learning~\cite{imran2015processing}. Due to the high variability of the data during a crisis, adapting the model to changes in features and their importance manually is undesirable (and often infeasible).

To overcome these issues, we use Deep Neural Networks (DNNs) to classify the tweets. DNNs are usually trained using online learning and have the flexibility to adaptively learn the model parameters as new batches of labeled data arrive, without requiring to retrain the model from scratch. DNNs use distributed condensed representation of words and learn the representation as well as higher level abstract features automatically for the classification task. Distributed representation (as opposed to sparse discrete representation)  generalizes well. This can be a crucial advantage at the beginning of a new disaster, when there is not enough event-specific labeled data. We can train a reasonably good DNN model using previously labeled data from other events, and then 
the model is fine-tuned adaptively as newly labeled data arrives in small batches.

In this paper, we use Deep Neural Network (DNN) to address two types of information needs of response organizations: \Ni identifying informative tweets and \Nii classifying them into topical classes. DNNs use distributed representation of words and learn the representation as well as higher level features automatically for the classification task. We propose a new online algorithm based on stochastic gradient descent to train DNNs in an online fashion during disaster situations. Moreover, we make our source code publicly available for crisis computing community for further research at:~\url{https://github.com/CrisisNLP/deep-learning-for-big-crisis-data}

In the next section, we provide details regarding DNNs we use and the online learning algorithm. Section~\ref{sec:settings} describes datasets and online learning settings. In Section~\ref{sec:results}, we describe results of our models. Section~\ref{sec:relatedwork} presents related-work and we conclude our paper in Section~\ref{sec:conclusion}. 
\section{Deep Neural Network} \label{sec:dnn}

As argued before, deep neural networks (DNNs) can be quite effective in classifying tweets during a disaster situation because of their distributed representation of words and automatic feature learning capabilities. Furthermore, DNNs are usually trained using online algorithms, which nicely suits the needs of a crisis response situation.   

Our main hypothesis is that in order to effectively classify tweets, which are short and informal, a classification model should learn the \emph{key} features at different levels of abstraction. To this end, we use a Convolutional Neural Network (CNN), which has been shown to be effective for sentence-level classification tasks \cite{kim:2014:EMNLP2014}.  

\subsection{Convolutional Neural Network} \label{sec:cnn}

\begin{figure}[tb!]
\centering
\includegraphics[height=2.5in]{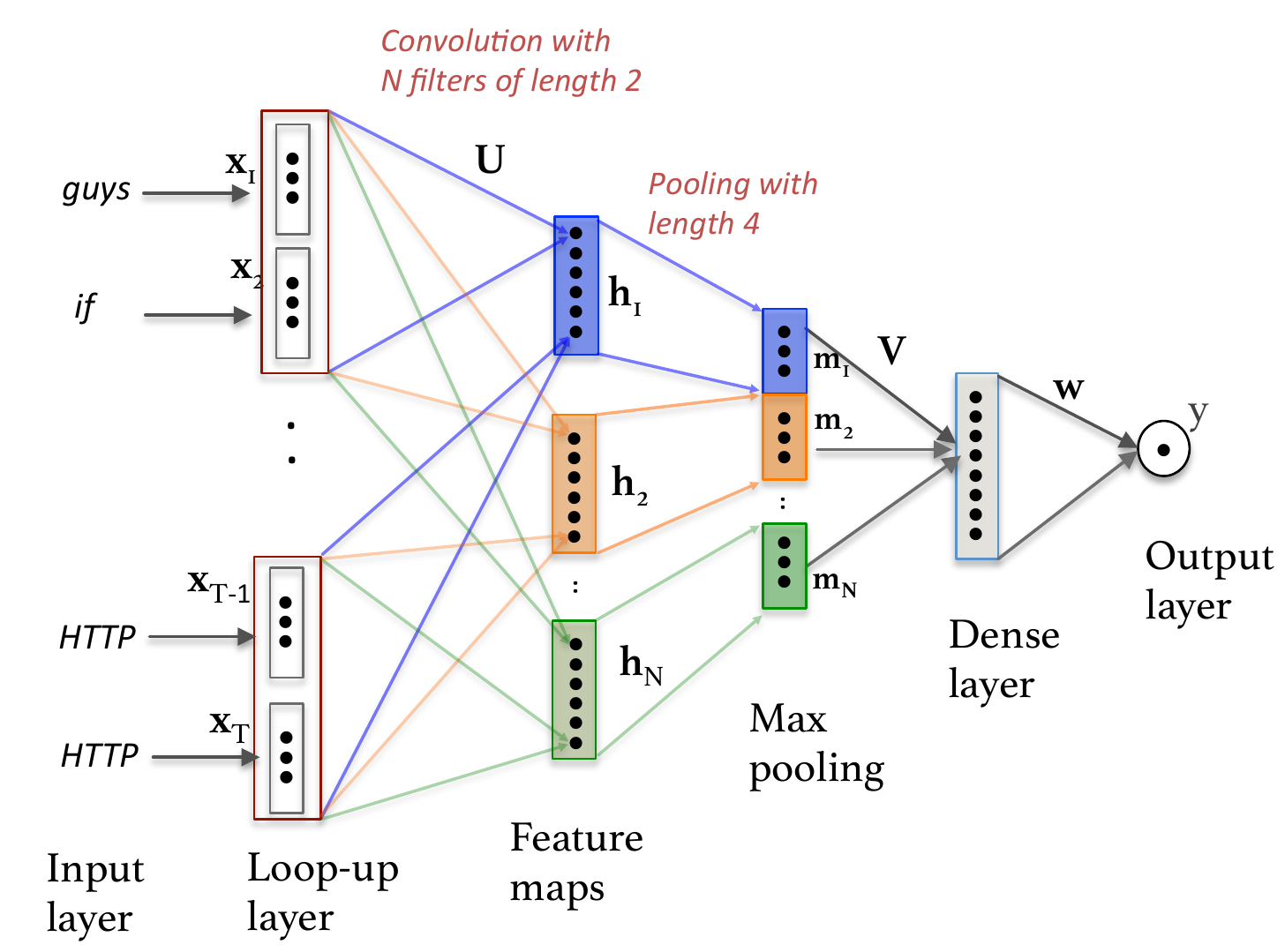}
\caption{Convolutional neural network on a sample tweet: ``guys if know any medical emergency around balaju area you can reach umesh HTTP doctor at HTTP HTTP''.}
\label{fig:cnn}
\end{figure}

Figure \ref{fig:cnn} demonstrates how 
a CNN works with an example tweet.\footnote{The HTTP tag in the example represents URLs.} Each word in the vocabulary $V$ is represented by a $D$ dimensional vector in a shared look-up table $L$ $\in$ $\real^{|V| \times D}$. $L$ is considered a model parameter to be learned. We can initialize $L$ randomly or
using pretrained word embedding vectors like word2vec \cite{mikolov2013efficient}.

Given an input tweet $\mathbf{s} = (w_1, \cdots, w_T)$, we first transform it into a feature sequence by mapping each word token $w_t \in \mathbf{s}$ to an index in $L$. The look-up layer then creates an input vector $\mathbf{x_t}\in\real^{D}$ for each token $w_t$, which are passed through a sequence of convolution and pooling operations to learn high-level abstract features. 

A convolution operation involves applying a \emph{filter} $\mathbf{u} \in \real^{L.D}$ to a window of $L$ words to produce a new feature  


\begin{equation}
h_t = f(\mathbf{u} . \mathbf{x}_{t:t+L-1} + b_t)
\end{equation}

\noindent where $\mathbf{x}_{t:t+L-1}$ denotes the concatenation of $L$ input vectors, $b_t$ is a bias term, and $f$ is a nonlinear activation function (e.g., $\sig, \tanh$). A filter is also known as a kernel or a feature detector. We apply this filter to each possible $L$-word window in the tweet to generate a \emph{feature map} $\mathbf{h}_i = [h_1, \cdots, h_{T+L-1}]$. We repeat this process $N$ times with $N$ different filters to get $N$ different feature maps. We use a \emph{wide} convolution \cite{Kalchbrenner14} (as opposed to \emph{narrow}), which ensures that the filters reach the entire sentence, including the boundary words. This is done by performing \emph{zero-padding}, where out-of-range (i.e., $t$$<$$1$ or $t$$>$$T$) vectors are assumed to be zero.  

After the convolution, we apply a max-pooling operation to each feature map.

\begin{equation}
\mathbf{m} = [\mu_p(\mathbf{h}_1), \cdots, \mu_p(\mathbf{h}_N)] \label{max_pool}
\end{equation}
  
\noindent where $\mu_p(\mathbf{h}_i)$ refers to the $\max$ operation applied to each window of $p$ features in the feature map $\mathbf{h}_i$. For instance, with $p=2$, this pooling gives the same number of features as in the feature map (because of the zero-padding).
Intuitively, the filters compose local $n$-grams into higher-level representations in the feature maps, and max-pooling reduces the output dimensionality while keeping the most important aspects from each feature map.


Since each convolution-pooling operation is performed independently, the features extracted become invariant in locations (i.e., where they occur in the tweet), thus acting like bag-of-$n$-grams. However, keeping the \emph{order} information could be important for modeling sentences. In order to model interactions between the features picked up by the filters and the pooling, we include a \emph{dense} layer of hidden nodes on top of the pooling layer 

\begin{equation}
\mathbf{z} = f(V\mathbf{m} + \mathbf{b_h}) \label{dense} 
\end{equation}

\noindent where $V$ is the weight matrix, $\mathbf{b_h}$ is a bias vector, and $f$ is a non-linear activation. The dense layer naturally deals with variable sentence lengths by producing fixed size output vectors $\mathbf{z}$, which are fed to the output layer for classification. 

Depending on the classification tasks, the output layer defines a probability distribution. For binary classification tasks, it defines a Bernoulli distribution:  

\begin{equation}
p(y|\mathbf{s}, \theta)= \Ber(y| \sig(\mathbf{w^T} \mathbf{z} + b )) \label{loss}
\end{equation}

\noindent  where $\sig$ refers to the sigmoid function, and $\mathbf{w}$ are the weights from the dense layer to the output layer and $b$ is a bias term. For multi-class classification the output layer uses a \texttt{softmax} function. Formally, the probability of $k$-th label in the output for classification into $K$ classes:


\begin{equation}
P(y = k|\mathbf{s}, \theta) = \frac{exp~(\mathbf{w}_k^T\mathbf{z} + b_k)} {\sum_{j=1}^{K} exp~({\mathbf{w}_j^T\mathbf{z} + b_j)}} \label{softmax}
\end{equation}

\noindent where, $\mathbf{w}_k$ are the weights associated with class $k$ in the output layer. We fit the models by minimizing the cross-entropy between the predicted distributions $\hat{y}_{n\theta} = p(y_n|\mathbf{s}_n, \theta)$ and the target distributions $y_n$ (i.e., the gold labels).\footnote{Other loss functions (e.g., hinge) yielded similar results.} The objective function $f(\theta)$ can be written as:


\begin{equation}
f (\theta) = \sum_{n=1}^{N} \sum_{k=1}^{K} y_{nk}~log~P(y_n = k|\mathbf{s}_n, \theta) \label{logloss}
\end{equation}

\noindent where, $N$ is the number of training examples and $y_{nk}$ $=$ $I(y_n = k)$ is an indicator variable to encode the gold labels, i.e., $y_{tk}=1$ if the gold label $y_t=k$, otherwise $0$.

\subsection{Online Learning} \label{sec:cnn_online}

DNNs are usually trained with first-order online methods like stochastic gradient descent (SGD). This
method yields a crucial advantage in crisis situations, where retraining the whole model each time a small batch of labeled data arrives is impractical. Algorithm \ref{alg:CNN} demonstrates how our CNN model can be trained in a purely online setting. We first initialize the model parameters $\theta_0$ (line 1), which can be a trained model from other disaster events or it can be initialized randomly to start from scratch.

As a new batch of labeled tweets $B_t= \{\mathbf{s}_1 \ldots \mathbf{s}_n \}$ arrives, we first compute the log-loss (cross entropy) in Equation \ref{logloss} for $B_t$ with respect to the current parameters $\theta_t$ (line 2a). Then, we use backpropagation to compute the gradients $f'(\theta_{t})$ of the loss with respect to the current parameters (line 2b). Finally, we update the parameters with 
the learning rate $\eta_t$ and 
the mean of the gradients (line 2c). We take 
the mean of the gradients to deal with minibatches of different sizes. Notice that we take only the current minibatch into account to get an updated model.  Choosing a proper learning rate $\eta_t$ can be difficult in practice. Several adaptive methods such as ADADELTA \cite{Zeiler12}, ADAM \cite{KingmaB14}, etc., have been proposed to overcome this issue. In our model, we use ADADELTA.

\small
\begin{algorithm}[t]
\SetAlgoLined
\hspace{-0.4cm} 1. Initialize the model parameters $\theta_0$; \\
\hspace{-0.4cm} 2. \For {a minibatch $B_t= \{\mathbf{s}_1 \ldots \mathbf{s}_n \}$ at time $t$} { 
  a. Compute the loss $f(\theta_{t})$ in Equation \ref{logloss}; \\
  b. Compute gradients of the loss $f'(\theta_{t})$ using backpropagation; \\
  c. Update: $\theta_{t+1} = \theta_{t} - \eta_t \frac{1}{n} f'(\theta_{t})$; \\
 }
\caption{Online learning of CNN}
\label{alg:CNN}
\end{algorithm}
\normalsize




\subsection{Word Embedding and Fine-tuning} \label{sec:word_emb}

As mentioned before, we can initialize the word embeddings $L$ randomly, and learn them as part of model parameters by backpropagating the errors to the look-up layer.
Random initialization may lead the training algorithm to get stuck in
a local minima. One 
can plug the readily available embeddings from external sources (e.g., Google embeddings \cite{mikolov2013efficient}) in the neural network model and use them as features without further task-specific tuning.
However, the latter
approach does not exploit the automatic feature learning capability of DNN models, which is one of the main motivations of using them. In our work, we use pre-trained word embeddings (see below) to better initialize our models, and we fine-tune them for our task, which turns out to be beneficial. 





Mikolov et al. \cite{mikolov2013efficient}  propose two log-linear models for computing word embeddings from large (unlabeled) corpuses efficiently: \Ni a \emph{bag-of-words} model CBOW that predicts the current word based on the context words, and \Nii a \emph{skip-gram} model that predicts surrounding words given the current word.\footnote{https://code.google.com/p/word2vec/} 
They released their pre-trained $300$-dimensional word embeddings trained by the skip-gram model on 
a Google news dataset.

Since we work on disaster related tweets, which are quite different from news, we have trained \emph{domain-specific} embeddings of $300$-dimensions (vocabulary size $20$ million) using the Skip-gram model of \textit{word2vec} tool \cite{mikolov2013distributed} from a large corpus of disaster related tweets. The corpus contains $57,908$ tweets and $9.4$ million tokens. 


\section{Dataset and Experimental\\ Settings}
\label{sec:settings}

In this section, we describe the datasets used for the classification tasks and the settings for CNN and online learning.


\subsection{Dataset and Preprocessing}
We use CrisisNLP~\cite{Imran2016CrisisNLP} labeled datasets. 
The CNN models were trained online using a labeled dataset related to the 2015 Nepal Earthquake\footnote{\url{https://en.wikipedia.org/wiki/April_2015_Nepal_earthquake}} 
and the rest of the datasets are used to train an initial model ($\theta_0$ in Algorithm \ref{alg:CNN}) upon which the online learning is performed. The Nepal earthquake dataset consists of approximately 12k labeled tweets collected from Twitter during the event using different keywords like \textit{NepalEarthquake}. Of all the labeled tweets, 9k are labeled by trained volunteers\footnote{These are trained volunteers from the Stand-By-Task-Force organization (http://blog.standbytaskforce.com/).} during the actual event using the AIDR platform~\cite{imran2014aidr} and the remaining 3k tweets are labeled using the Crowdflower\footnote{\url{crowdflower.com}} crowdsourcing platform. 

The dataset is labeled into different informative classes (e.g., affected individuals, infrastructure damage, donations etc.) and one ``not-related'' or ``irrelevant'' class. Table \ref{tbl:classes} provides a one line description of each class and also the total number of labels in each class. \emph{Other useful information} and \emph{Not related or irrelevant} are the most frequent classes in the dataset. 

\begin{table*}[htb]
\centering
\footnotesize
\caption{Description of the classes in the dataset. Column \emph{Labels} shows the total number of labeled examples in each class}
\label{tbl:classes}
\begin{tabular}{lrl}
\toprule
\textbf{Class} & \textbf{Labels} & \textbf{Description}  \\
\midrule
Affected individuals  & 756 & \begin{tabular}[c]{@{}l@{}}Reports of deaths, injuries, missing, found, or displaced people\end{tabular} \\
\begin{tabular}[c]{@{}l@{}}Donations and volunteering\end{tabular}  & 1021 & \begin{tabular}[c]{@{}l@{}}Messages containing donations (food, shelter, services etc.) or volunteering offers\end{tabular} \\
\begin{tabular}[c]{@{}l@{}}Infrastructure and utilities\end{tabular} & 351 & Reports of infrastructure and utilities damage \\
Sympathy and support & 983 & Messages of sympathy-emotional support \\
Other useful information & 1505 & \begin{tabular}[c]{@{}l@{}}Messages containing useful information that does not fit in one of the above classes\end{tabular} \\
Not related or irrelevant & 6698 & \begin{tabular}[c]{@{}l@{}}Irrelevant or not informative, or not useful for crisis response\end{tabular}    \\
\bottomrule                                 
\end{tabular}
\end{table*}

\smallskip
\noindent{\bf Data Preprocessing:} We normalize all characters to their lower-cased forms, truncate elongations to two characters, spell out every digit to \texttt{D}, all twitter usernames to \texttt{userID}, and all URLs to \texttt{HTTP}. We remove all punctuation marks except periods, semicolons, question and exclamation marks. We further tokenize the tweets using the CMU TweetNLP tool \cite{gimpel2011part}.


\subsection{Online Training Settings}

Before performing the online learning, we assume that an initial model $\theta_0$ exists. In our case, we train the initial model using all the datasets from CrisisNLP except the Nepal earthquake. For online training, we sort the Nepal labeled data based on the time stamp of the tweets. This brings the tweets in their posting order. Next, the dataset $D$ is divided at each time interval $d_t$ in which case $D$ is defined as: D = $\sum_{t=1}^T d_t$ where $d_t= 200$.
For each time interval $t$, we divide the available labeled dataset into a train set (70\%), dev set (10\%), and a test set (20\%) 
%
using ski-learn toolkit's module \cite{scikit-learn}, which ensured that the class distribution remains reasonably balanced in each subset.



Based on the data splitting strategy mentioned above, we start online learning to train a binary and a multi-class classifier. For the binary classifier training, we merge all the informative classes to create one general \emph{Informative} class. We train CNN models by optimizing the cross entropy in Equation \ref{loss} using the gradient-based online learning algorithm ADADELTA \cite{Zeiler12}.\footnote{Other algorithms (SGD, Adagrad) gave similar results.} The learning rate and
the parameters were set to the values as suggested by the authors. 
The maximum number of epochs was set to $25$. To avoid overfitting, we use dropout \cite{Srivastava14a} of hidden units and \emph{early stopping} based on the accuracy on the validation set.\footnote{$l_1$ and $l_2$ regularization on weights did not work well.} We experimented with $\{0.0, 0.2, 0.4, 0.5\}$ dropout rates and $\{32, 64, 128\}$ minibatch sizes. 
We limit the vocabulary ($V$) 
to the most frequent $P\%$ ($P\in\{80, 85, 90\}$) words in the training corpus. The word vectors in $L$ were initialized with the pre-trained embeddings.
We use rectified linear units (ReLU) for the activation functions ($f$), $\{100, 150, 200\}$ filters each having window size ($L$) of $\{2, 3, 4\}$, pooling length ($p$) of $\{2,3, 4\}$, and $\{100, 150, 200\}$ dense layer units. All the hyperparameters are tuned on the development set.


\section{Results}
\label{sec:results}

In this section, we present our results for binary and multi-class classification tasks.

\subsection{Binary Classification}
%

\begin{figure}[h]
\begin{center}
\begin{tikzpicture}
\begin{axis}[
  xlabel= Train,
  ylabel= AUC]
\addplot table {binary.dat};
\end{axis}
\end{tikzpicture}
\caption{\label{fig:online_bin} {\bf Binary} classification: Performance of the CNN model with varying size of the training data}
\end{center}
\end{figure}
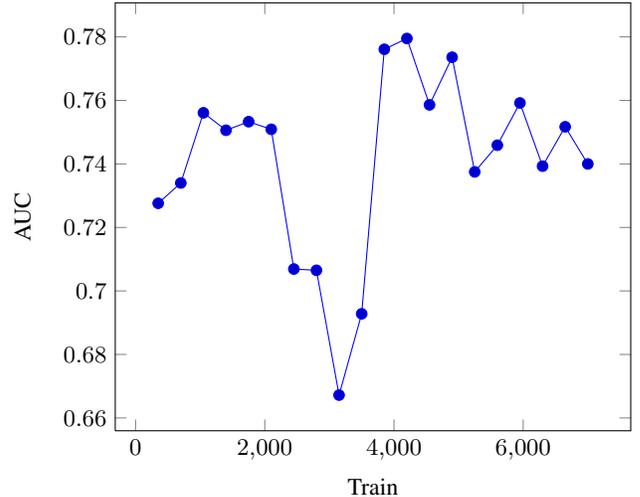

Figure \ref{fig:online_bin} shows the results for 
the \textit{``informative"} vs. \textit{``not informative"} binary classification task using online learning. 
The performance of the model is quite inconsistent as the size of the in-event training data varies. We observe an improvement in performance initially. However, the results dropped when the training size is between 2200 to 3900 tweets. We investigated this strange result and found that this could be due to the inconsistencies in the annotation procedure and the data sources. In our in-event (Nepal Earthquake) training data, first 3000 tweets are from CrowdFlower and the rest are from AIDR.  Tweets in CrowdFlower were annotated by paid workers, where AIDR tweets are annotated by volunteers. We speculate these inconsistencies can affect the performance at the beginning, but as the model sees more AIDR data (4000+), the performance stabilizes.




%

\subsection{Multi-Class Classification}

\begin{figure}[h]
\begin{center}
\begin{tikzpicture}
\begin{axis}[
  xlabel= Train,
  ylabel= Accuracy ]
\addplot table {multil.dat};
\end{axis}
\end{tikzpicture}
\caption{\label{fig:online_mul} {\bf Multi-class} classification: Performance of the CNN model with various sizes of training data.}
\end{center}
\end{figure}
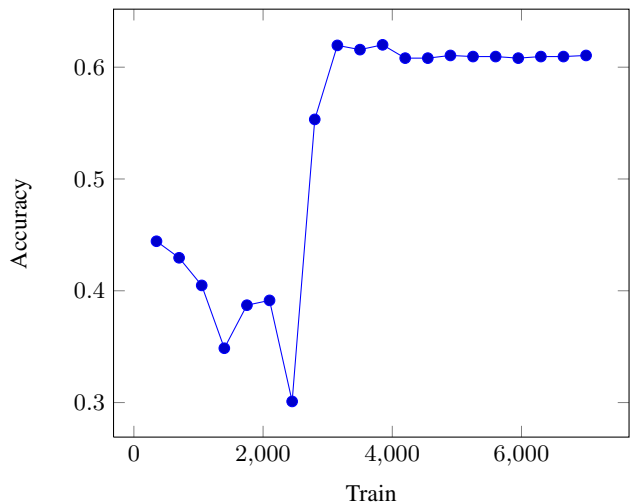

Figure \ref{fig:online_mul} summarizes the results of online training for
the multi-class classification task. Since multi-class classification is a harder task than binary classification, the first training run provides very low accuracy and the results continue to drop until
a good number of training examples are available, which in this case is approximately 2200 labeled tweets. 
As in the binary classification case,
after the initial dip in performance, once over 3000 tweets are available, 
the performance of the classifier 
improves and remains stable after that.

The benefit of using online learning methods like CNN compared to offline learning methods used in classifiers like SVM, Naive Bayes, and Logistic Regression is online training. The labeled data comes in batches and retraining a model on the complete data every time with the addition of newly
labeled data is an expensive task. Online training methods learn in small batches, which suits the situation in hand perfectly.

Another advantage of neural network methods is automatic feature extraction that does not require any manual feature engineering. The models take labeled tweets as input and automatically learn features based on distributed representation of words.

\subsection{Discussion}
Rapid analysis of social media posts during time-critical situations is important for humanitarian response organization to take timely decisions and to launch relief efforts. This work proposes solutions to two main challenges that humanitarian organizations face while incorporating social media data into crisis response. First, how to filter-out noisy and irrelevant messages from big crisis data and second, categorization of the informative messages into different classes of interest. By utilizing labeled data from past crises, we show the performance of DNNs trained using the proposed online learning algorithm for binary and multi-class classification tasks. 

We observe that past labeled data helps when no event-specific data is available in the early hours of a crisis. However, labeled data from event always help improve the classification accuracy.

\section{Related work}
\label{sec:relatedwork}
Recent studies have shown the usefulness of crisis-related data on social media for disaster response and management~\cite{acar2011twitter,sakaki2010earthquake,varga2013aid}.
A number of systems have been developed to classify, extract, and summarize~\cite{rudra2016summarizing} crisis-relevant information from social media; for a detailed survey see~\cite{imran2015processing}. Cameron, et al., describe a platform for emergency situation awareness~\cite{cameron2012emergency}. 
They classify interesting tweets using an SVM classifier. 
Verma, et al., use Naive Bayes and MaxEnt classifiers to find situational awareness tweets from several crises~\cite{verma2011natural}. 
Imran, et al., implemented AIDR to classify a Twitter data stream during crises~\cite{imran2014aidr}. They use a random forest classifier in an offline setting. 
After receiving every mini-batch of 50 training examples, 
they replace the older model with a new one. In~\cite{imran2016cross}, the authors show the performance of a number of non-neural network classifiers trained on labeled data from past crisis events. However, they do not use DNNs in their comparison.
 


DNNs and word embeddings have been applied successfully to address NLP problems~\cite{collobert2011natural,carageaidentifying,nguyen2016rapid,liu-joty-meng:2015:EMNLP,Joty-hoque-acl16}. The emergence of tools such as word2vec \cite{mikolov2013distributed} and GloVe \cite{pennington-socher-manning:2014:EMNLP2014} have enabled NLP researchers to learn word embeddings efficiently and use them to train better models. 

Collobert, et al. \cite{collobert2011natural} presented a unified DNN architecture for solving various NLP tasks including part-of-speech tagging, chunking, named entity recognition and semantic role labeling. They showed that DNNs outperform traditional models in most of these
tasks.  They also proposed a multi-task learning framework for solving the tasks jointly. 

Kim \cite{kim:2014:EMNLP2014} and Kalchbrenner et al. \cite{Kalchbrenner14} used convolutional neural networks (CNN) for sentence-level classification tasks (e.g., sentiment/polarity classification, question classification) and showed that CNNs outperform traditional methods (e.g., SVMs, MaxEnts). Caragea, Silvescu, and Tapia used CNNs to identify informative messages during disasters~\cite{carageaidentifying}. 
However, to the best of our knowledge, no previous research has shown the efficacy of CNNs to both the binary classification and the multi-class classification problems using online learning.

\section{Conclusions}
\label{sec:conclusion}
We presented an online learning model namely Convolutional Neural Network for the purpose of classifying tweets in a disaster response scenario. We proposed a new online learning algorithm for training CNNs in online fashion. We showed that online training of the model perfectly suits the disaster response situation. We assume that a base model trained on past crisis labeled data exists and the event-specific labeled data arrive in small batches which are used to perform online learning. The neural network models bring an additive advantage of automatic feature extraction which eases the training process when compared with offline learning methods like SVM, logistic regression. The model uses only labeled tweets for training and automatically learns features from them. We reported the results of two classification tasks (i.e. binary and multi-class). Moreover, we also provide source code for the online learning of CNN models to research community for further extensions.


\bibliographystyle{abbrv}

\bibliography{sigproc}
\balance
\end{document}

%% file: defs.tex
\DeclareMathOperator \real{\mathbb{R}}

\DeclareMathOperator*{\Ber}{Ber}
\DeclareMathOperator*{\sig}{sig}

\newcommand{\Ni}{({\em i})~}
\newcommand{\Nii}{({\em ii})~}